\documentclass[conference]{IEEEtran}
\newcommand{\myheader}{Preprint submitted as a conference paper at the \textit{IEEE International Conference on Developmental Learning and Epigenetic Robotics (ICDL-Epirob)} 2019}

\usepackage{atbegshi,picture}

\usepackage{datetime}            

\usepackage{lastpage}            

\newcommand{\myleftstd}{1.5cm}

\ifthenelse{\isodd{\thepage}}
 {\newcommand{\myleftmargin}{\oddsidemargin+\myleftstd}}
 {\newcommand{\myleftmargin}{\evensidemargin+\myleftstd}}

\AtBeginShipoutNext { 
 \AtBeginShipoutUpperLeft{%
  \put(\dimexpr 0.0\myleftmargin\relax,-1.2cm){\parbox{\textwidth}{\footnotesize\sffamily\noindent{\centering \myheader \hfill}}}%
}}

\usepackage{amsmath} 
\usepackage{amssymb}  
\usepackage{algorithm}
\usepackage{algorithmicx}
\usepackage[noend]{algpseudocode}

\algrenewcommand\alglinenumber[1]{
    {\sf\footnotesize\addfontfeatures{Colour=888888,Numbers=Monospaced}#1}}
\algrenewcommand\algorithmicrequire{\textbf{Precondition:}}
\algrenewcommand\algorithmicensure{\textbf{Postcondition:}}

\usepackage{todonotes}

\usepackage{graphicx}
\usepackage{xcolor}
\colorlet{darkblue}{blue!50!black}
\colorlet{hlinkcolor}{darkblue}
\usepackage[colorlinks=true,linkcolor=hlinkcolor,urlcolor=hlinkcolor,citecolor=hlinkcolor]{hyperref}

\makeatletter
\let\NAT@parse\undefined
\makeatother
\usepackage[numbers]{natbib}

\renewcommand{\todo}[1]{}

\def\BibTeX{{\rm B\kern-.05em{\sc i\kern-.025em b}\kern-.08em
    T\kern-.1667em\lower.7ex\hbox{E}\kern-.125emX}}

\begin{document}

\title{
Curriculum goal masking for continuous deep reinforcement learning \\
\thanks{The authors gratefully acknowledge support from the German Research Foundation DFG under project CML (TRR 169), the Volkswagen Stiftung and the NVIDIA corporation.}
}

\author{\IEEEauthorblockN{1\textsuperscript{st} Manfred Eppe}
\IEEEauthorblockA{\textit{Department of Informatics} \\
\textit{ Universit\"at Hamburg}\\
Germany \\
eppe@informatik.uni-hamburg.de}
\and
\IEEEauthorblockN{2\textsuperscript{nd} Sven Magg}
\IEEEauthorblockA{\textit{Department of Informatics} \\
\textit{ Universit\"at Hamburg}\\
Germany \\
magg@informatik.uni-hamburg.de}
\and
\IEEEauthorblockN{3\textsuperscript{rd} Stefan Wermter}
\IEEEauthorblockA{\textit{Department of Informatics} \\
\textit{ Universit\"at Hamburg}\\
Germany \\
wermter@informatik.uni-hamburg.de}
}

\maketitle
\thispagestyle{empty}
\pagestyle{empty}

\begin{abstract}
	Deep reinforcement learning has recently gained a focus on problems where policy or value functions are independent of goals. Evidence exists that the sampling of goals has a strong effect on the learning performance, but there is a lack of general mechanisms that focus on optimizing the goal sampling process. In this work, we introduce goal masking as a method to estimate a goal's difficulty level and to exploit this estimation to realize curriculum learning. 
Our results indicate that focusing on goals with a medium difficulty level is appropriate for deep deterministic policy gradient (DDPG) methods, while an ``aim for the stars and reach the moon-strategy'', where difficult goals are sampled much more often than simple goals, leads to the best learning performance in cases where DDPG is combined with hindsight experience replay (HER).
\end{abstract}

\section{INTRODUCTION}
\label{sec:intro}
Human infants have been reported to learn according to the Goldilocks principle, where they choose learning goals that have just the right difficulty level to maximize the learning performance (e.g. \cite{Kidd2014}). 
Such Goldilocks goals are involved in curriculum learning (CL), (e.g. \cite{Bengio2009}), where the learner does not attempt to solve a hard problem immediately using a brute-force trial-and-error strategy but instead attempts to follow a certain curriculum that involves a favorable sequence of subgoals within the learning process.  
The CL approach implies that, as an agent learns to master simple tasks, it will, at some point, not learn from such cases anymore and must try to accomplish more challenging goals. Ideally, the agent will train under Goldilocks conditions, where goals are neither too hard nor too easy to achieve. 

As we summarize in Sec.~\ref{sec:related}, the CL approach has been investigated also computationally and in robotic contexts (e.g.\cite{Bengio2009,Pentina2015,Matiisen2017}). However, the computational perspective on CL yields several non-trivial problems. 
The problem we address in this work is that it is hard to estimate the difficulty of a goal during the training process because of two reasons: Firstly, in a continuous state space, the difficulty of a previously unseen goal cannot be simply interpolated from the difficulty level of similar goals that have been tackled before. 
Secondly, the goal difficulty is dynamic in the sense that goals will become easier to achieve as the agent learns. 
These two issues motivate our following central research question:
\begin{center}
\textit{
How and under which assumptions can we dynamically estimate the difficulty level of continuous reinforcement learning goals, and how can we efficiently sample goals with a certain desired difficulty level to maximize the learning performance?}
\end{center}

In addressing these questions, we build on recent advances on universal value function approximators (UVFA) \cite{Schaul2015} which allow for algorithms that are capable of learning policies that can be conditioned to goals. 
%

As our core contribution, we propose a general and efficient \emph{curriculum goal masking } (CGM) method to automatically create goals of appropriate difficulty (see Sec.~\ref{sec:cgm}). 
The masking allows for estimating the difficulty of a previously unseen masked goal by considering the past success rate of the learner for goals to which the same mask has been applied. In our evaluation (Sec.~\ref{sec:exp_results}), we determine the optimal Goldilocks difficulty level to maximize the learning performance for a combination of two different RL algorithms, namely deep deterministic policy gradient (DDPG) \cite{Lillicrap2016} and hindsight experience replay (HER) \cite{Andrychowicz2017}. Both are general state-of-the-art solutions to address continuous goal-independent deep reinforcement learning. We perform the evaluation for two robotic object manipulation tasks and conclude in Sec.~\ref{sec:conclusion} that extending the investigated algorithms with CGM provides a significant performance boost, especially for the more difficult pick-and-place problem that involves the implicit learning of causal relations.


\begin{figure*}[ht]
\centering
\includegraphics[width=.24\textwidth,trim={20cm 13cm 25cm 8cm},clip]{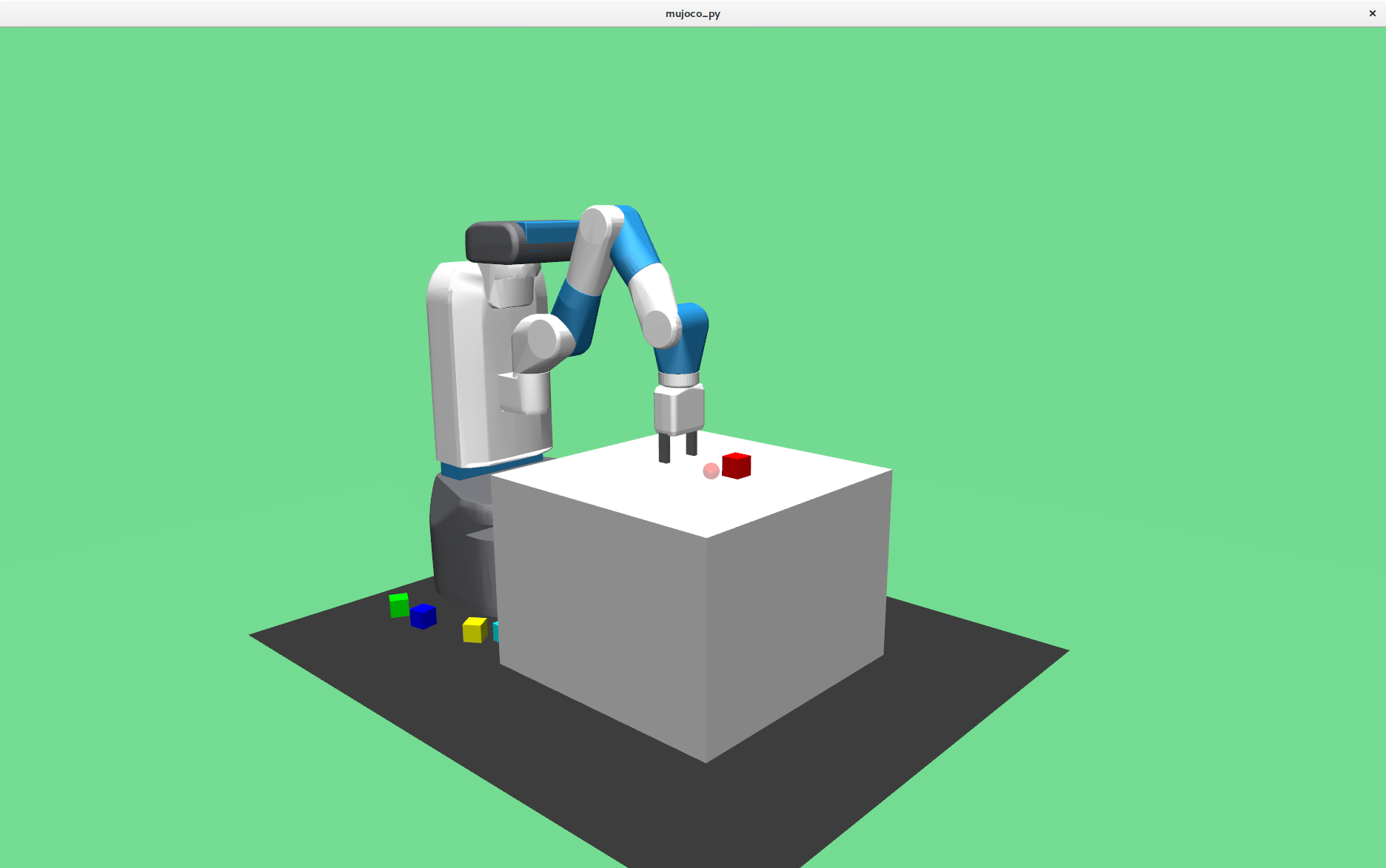}
\hspace{-0.75cm}\begin{minipage}[b][3.9cm][t]{0.7cm}(a)\end{minipage} 
\includegraphics[width=.24\textwidth,trim={20cm 13cm 25cm 8cm},clip]{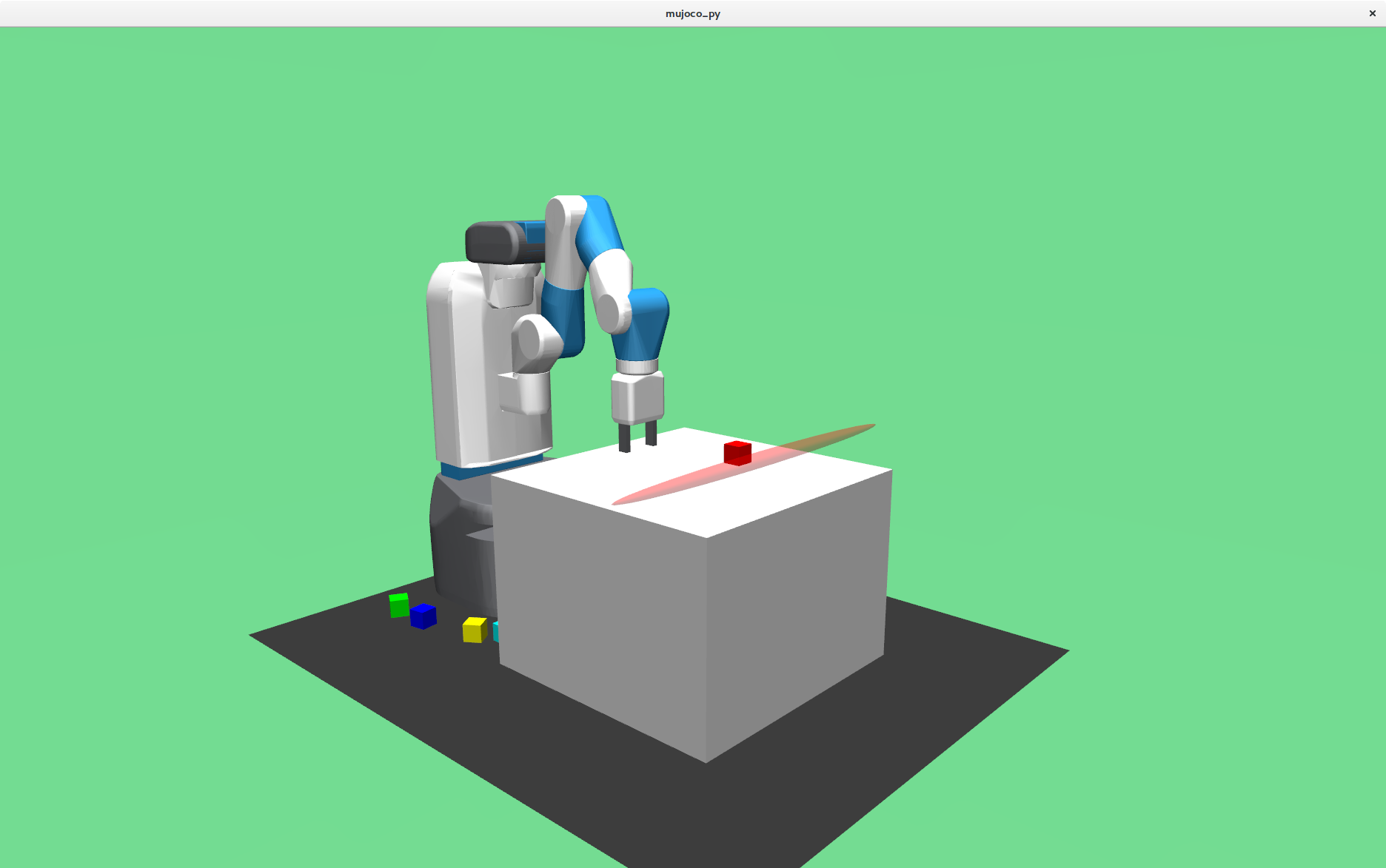}
\hspace{-0.75cm}\begin{minipage}[b][3.9cm][t]{0.7cm}(b)\end{minipage} 
\includegraphics[width=.24\textwidth,trim={20cm 13cm 25cm 8cm},clip]{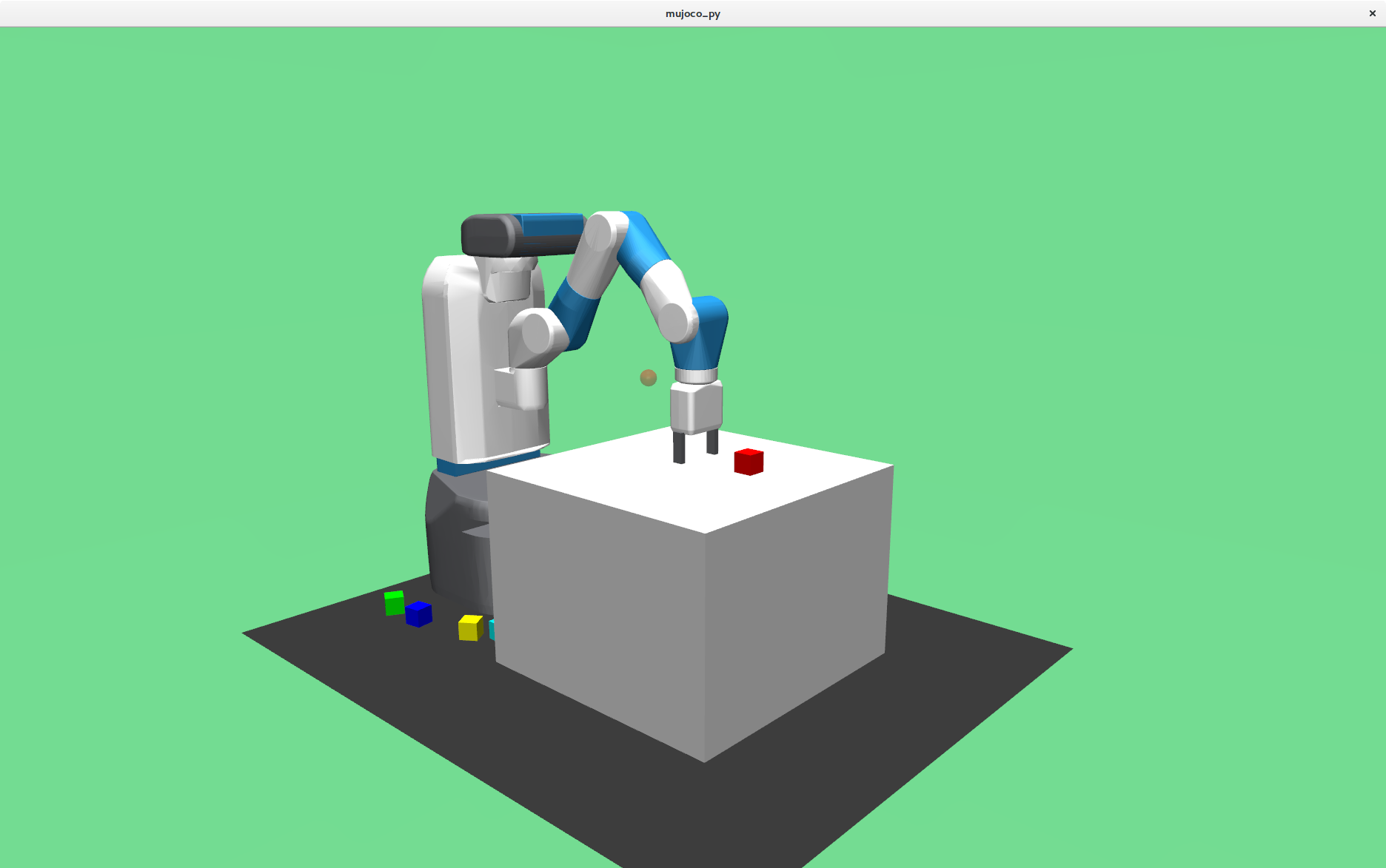}
\hspace{-0.75cm}\begin{minipage}[b][3.9cm][t]{0.7cm}(c)\end{minipage} 
\includegraphics[width=.24\textwidth,trim={20cm 13cm 25cm 8cm},clip]{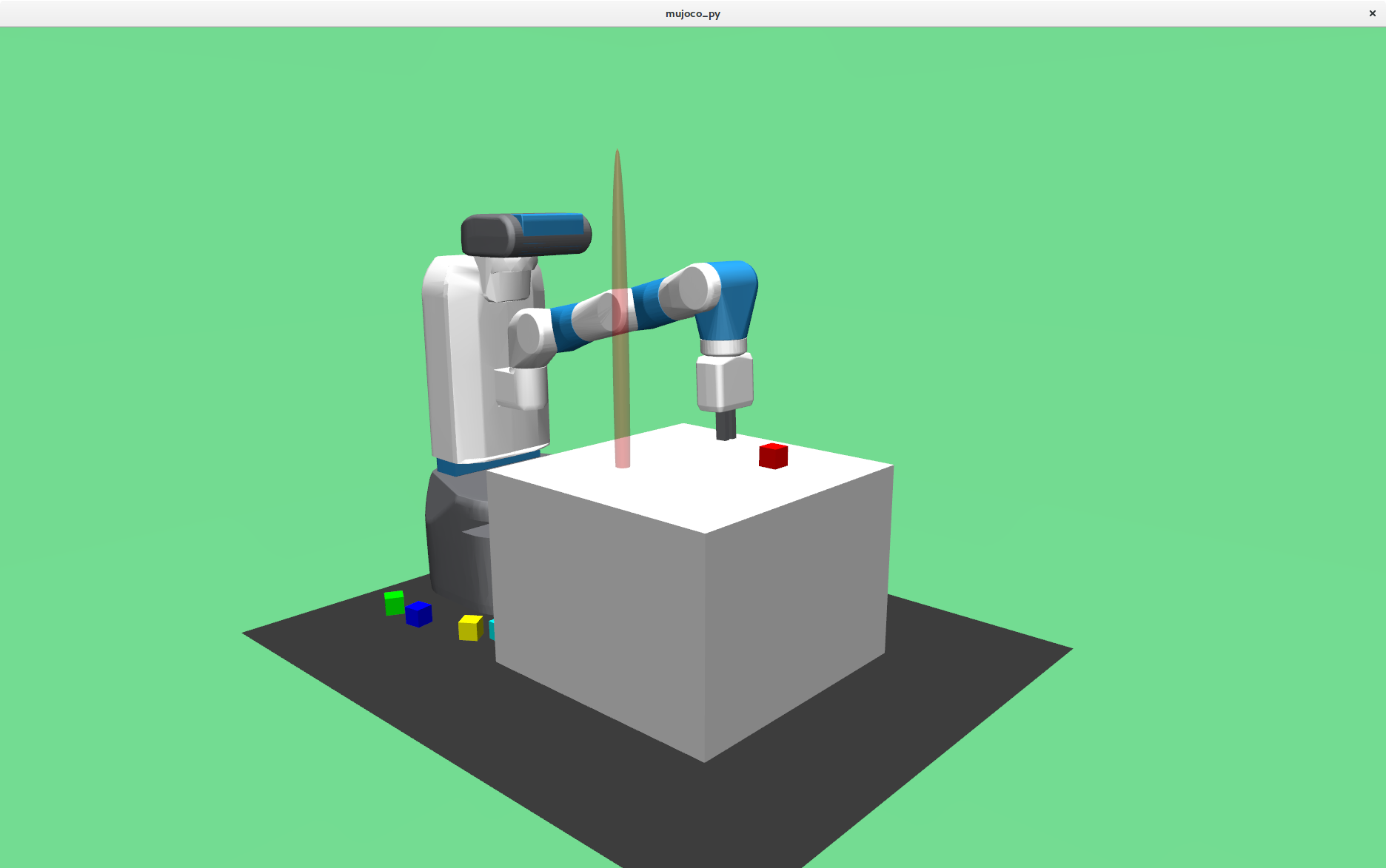}
\hspace{-0.75cm}\begin{minipage}[b][3.9cm][t]{0.7cm}(d)\end{minipage} 
\vspace{-.2cm}
\caption{The experimental setup. The robot must move the red block to the target indicated by the red sphere. (a) Push task: the target is on the table's surface. (b) Push task: the target's x-coordinate is masked. (c) Pick and place task: the target is above the table's surface. (d) Pick and place task: the target's z-coordinate is masked. 
}
\vspace{-.2cm}
\label{fig:experiments}
\end{figure*}

\section{RELATED WORK}
\label{sec:related}
\subsection{Curriculum learning for deep reinforcement learning}
The term curriculum learning (CL) for neural networks was coined by \citet{Bengio2009},
\todo{Maybe this is too strong, Elman also used this strategy already, but just did nit name it CL, as far as I know... Maybe elaborate a bit on the historical development of the idea of CL.}
 describing the idea that solving easier problems first has advantages to learn more complex goals later and thus that learning can be optimized by presenting the problems in an optimal order, a curriculum. The problem of automatically finding this optimal order has been addressed by several researchers, e.g. \citet{Pentina2015} who propose to use a generalization bound criterion to order multiple tasks that have to be learned, \citet{Matiisen2017} who introduce a teacher-learner method where the teacher selects goals depending on the slope of the learning curve for those subtasks, or \citet{Graves2017} who introduce intrinsically motivated curriculum learning, using prediction gain as well as complexity gain measures to select subtasks. 
Generally, approaches using such a CL method show performance gains in the learning process when tasks are presented to the neural network in a favorable sequence, instead of learning all tasks in parallel. 
A limiting factor is that most of the existing approaches assume a given set of predefined subtasks, relying on previous knowledge, and often the proposed ordering process is computationally expensive, adding to the already costly learning process. 

Recently, CL has also been used for reinforcement learning scenarios. For example, \citet{Narvekar2016,Narvekar2017} have presented a method to automatically define subtasks for a given target task, as well as learning the optimal sequence by incorporating a parallel reinforcement loop. The generation of subgoals includes a method to simplify the problem, but the authors assume it to be predefined for a given problem domain. Our masking method constitutes an automated, domain-independent, and more general solution of such a simplification approach. 

A similar automated approach has recently been proposed by \citet{Florensa2018AutomaticAgents}. The authors introduce a  Generative Adversarial Network (GAN) that is trained to produce \emph{goals of intermediate difficulty} (GOID) which have the dynamic optimal difficulty given the current learning progress. For measuring the difficulty, the authors use the training success rate of the previous time steps for the same goal. In their experiments, \citeauthor{Florensa2018AutomaticAgents} combine their goal sampling method with the trust region policy optimization (TRPO) algorithm \cite{Schulman2015a}, which they also use as a baseline for their evaluation. 
In contrast to our work, the authors have not yet combined their algorithm with hindsight experience replay (HER) \cite{Andrychowicz2017}, but propose to do so in future work. Another difference to our work is that training a GAN to generate goals is computationally more expensive than our goal masking process. 
Interestingly, \citet{Florensa2017} show that curriculum learning is also feasible in the reverse direction, in the sense that the starting state has an effect on the difficulty of a task. They realize this idea of inverse curriculum generation by defining \emph{starts of intermediate difficulty} (SOID) which imply an optimal difficulty with respect to the learning goal.

A multi-agent implementation of automated curriculum generation is the work by \citet{Sukhbaatar2018IntrinsicSelf-Play} who propose to use two agents using intrinsically motivated self-play that generate goals for each other. Their results are comparable to those by \citet{Florensa2018AutomaticAgents}, but the authors do not evaluate their work in a robotic domain. 

Curriculum learning is also related to self-paced learning (SPL) \cite{Kumar2010}, where samples are increasingly difficult to solve. It thus is similar to CL in that it produces a kind of curriculum, but the sequence is not relying on a predefined set of goals. The complexity of samples is estimated by monitoring the training error and thus the current performance of the learner. This approach has also been combined with CL \cite{Jiang2015}, but despite the learner-centric influence of SPL, the approach still relies on a predefined set of goals ``predetermined by an oracle'' \cite{Jiang2015}. 

\citet{Fournier2018} and \citet{Kerzel2018} also present approaches to simplify goals. However, instead of masking the goals, they simplify tasks by reducing the required precision for successful completion. Although the accuracy measurement is similar and the task simplification is of comparable complexity to our goal masking method, our results show a better performance gain and are also evaluated on a more complex pick-and-place task. 

\subsection{Continuous goal-independent reinforcement learning}
Most reinforcement learning approaches build on manually defined reward functions based on a metric that is specific to a single goal, such as the body height, posture and forward speed of a robot that learns to walk \cite{Schulman2015a}. Goal-independent reinforcement learning settings do not require such reward shaping (cf. \cite{Ng1999}), as they allow one to parameterize learned policies and value functions with goals. At each step $t$, the agent executes an action $a_t \in \mathbb{R}^n$ given an observation $o_t \in \mathbb{R}^m$ and a goal $g \subseteq \mathbb{R}^m$, according to a policy $\pi$ that maps from the current observation and goal to a probability distribution over actions. 

An action generates a reward $r_t$ if the goal is achieved at time t. 
To decide whether a goal has been achieved, a function $f(o_t)$ is defined that maps the observation space to the goal space, and the goal is considered to be achieved if $|f(o_t) - g| < \epsilon$ for a small threshold $\epsilon$. This sharp distinction of whether or not a goal has been achieved based on a distance threshold causes the reward to be sparse and renders shaping the reward with a hand-coded reward function unnecessary.

The DDPG algorithm \cite{Lillicrap2016} is an off-policy actor-critic deep reinforcement learning method. For the goal-independent case, a rollout worker process performs a fixed number $n_r$ of $n_p$ parallel rollout episodes during each epoch and records a state transition $s=(o_t,g,a_t,o_{t+1})$ for each action step.
After each set of parallel rollouts, the neural network models for the actor and the critic are trained off-policy. 

Hindsight experience replay (HER) \cite{Andrychowicz2017} has been presented as an extension to all continuous goal-independent RL algorithms, including DDPG. In brief, HER considers unsuccessful rollouts as positive learning examples by pretending in hindsight that for a current step $t$, an observation $o_{t+l}$, that has been achieved at some random future step $t+l$, was actually the goal state. Formally, state transitions $(o_t,g,a_t,o_{t+1})$ in the replay buffer are modified after each rollout cycle, such that $g$ is replaced by $f(o_{t+l})$, where $l \in (0,T-t)$ is randomly chosen, with $T$ being the maximal number of steps of the respective rollout. \citet{Andrychowicz2017} demonstrate that the best learning performance is achieved if 80\% to 90\% of all state transition samples are modified this way.


\section{CURRICULUM GOAL MASKING (CGM)}
\label{sec:cgm}
In the introduction (Sec.~\ref{sec:intro}), we identify a major challenge for realizing CL-based RL methods, namely the problem of estimating the difficulty of a goal and to sample goals according to their dynamic difficulty level. 
As a solution, we propose to mask combinations of subgoals, and to associate a difficulty level with each mask. 
We consider a training rollout to be successful if those subgoals are achieved that are not masked. For example, if the z-axis of the goal of the pick-and-place task is masked, as in \autoref{fig:experiments} (d), the goal is considered to be achieved if only the x- and y coordinates are reached. 
Assuming that all goals created with the same mask share approximately the same difficulty allows for estimating the difficulty level of a simplified goal as the difficulty level assigned to the corresponding goal mask. 

Formally, given a goal vector $g \in \mathbb{R}^n$, we define the space of goal masks as $M = \{1,0\}^n$, and we apply a goal mask $m \in M$ as follows: Let $o_t$ be the current observation at a rollout or training step $t$. For that step, we mask the original goal $g$ to obtain the masked goal $g^m_t$ by keeping those elements where the mask is 1 and by setting all other elements to the corresponding values of the observation vector $o_t$. 
\begin{equation}
\label{eq:goal_mask}
g^m_t = g \odot m + o_t \odot (-m +1)
\end{equation}
This effectively renders those elements of the goal vector that are masked to be achieved at the current step, which causes the reward to be provided if all non-masked goal elements are achieved.\footnote{In preliminary experiments, we also tried to implement the masking by setting masked subgoals and their corresponding elements of the observation vector to zero, as in recent neural attention models (e.g. \cite{Bahdanau2015}), but this resulted in inferior performance.}
This masking approach allows us to address our second research question, by estimating a success chance for each subgoal combination as the average over the successes of the last $h$ evaluation rollouts for goals created by this mask. 

\begin{figure*}[ht!]
\centering
\includegraphics[width=.475\textwidth,trim={1.38cm 0.1cm 2.5cm 0.4cm},clip]{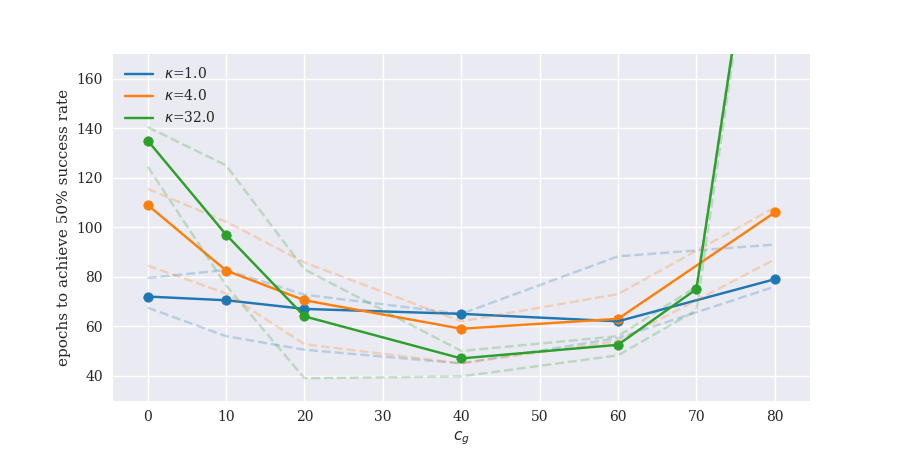}
\hspace{-2.7cm}\begin{minipage}[b][26pt][t]{2.4cm}\tiny (a) Push object DDPG\end{minipage} 
\hspace{0.3cm}
\includegraphics[width=.475\textwidth,trim={1.38cm 0.1cm 2.5cm 0.4cm},clip]{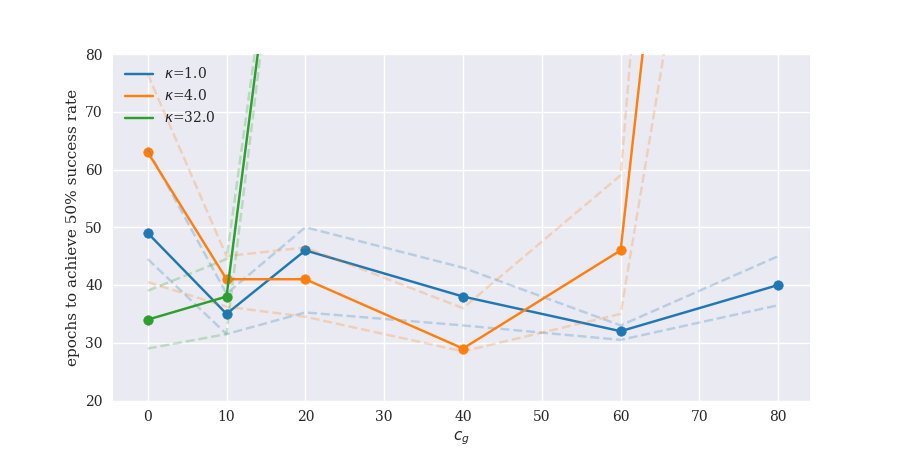}
\hspace{-2.7cm}\begin{minipage}[b][26pt][t]{2.4cm}\tiny (b) Pick-and-place  DDPG\end{minipage} 
\vspace{-0.3cm}
\includegraphics[width=.475\textwidth,trim={1.38cm 0.1cm 2.5cm 0.4cm},clip]{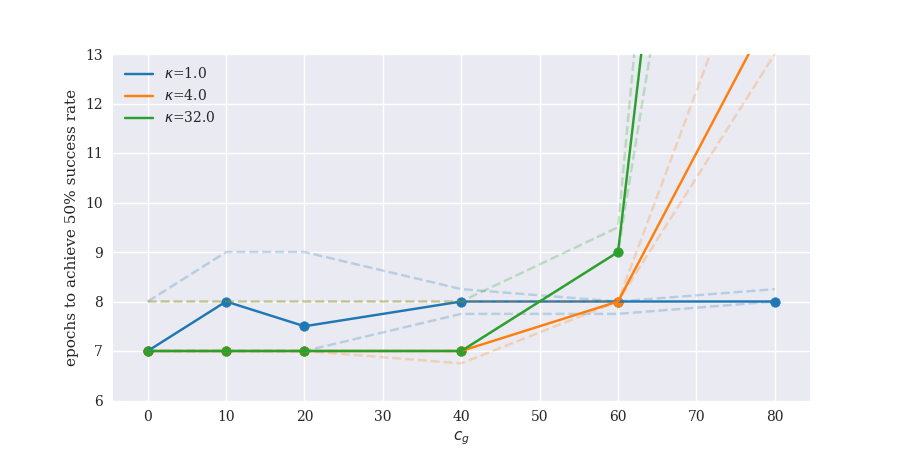}
\hspace{-2.7cm}\begin{minipage}[b][26pt][t]{2.4cm}\tiny (c) Push object DDPG+HER\end{minipage} 
\hspace{0.3cm}
\includegraphics[width=.475\textwidth,trim={1.38cm 0.1cm 2.5cm 0.4cm},clip]{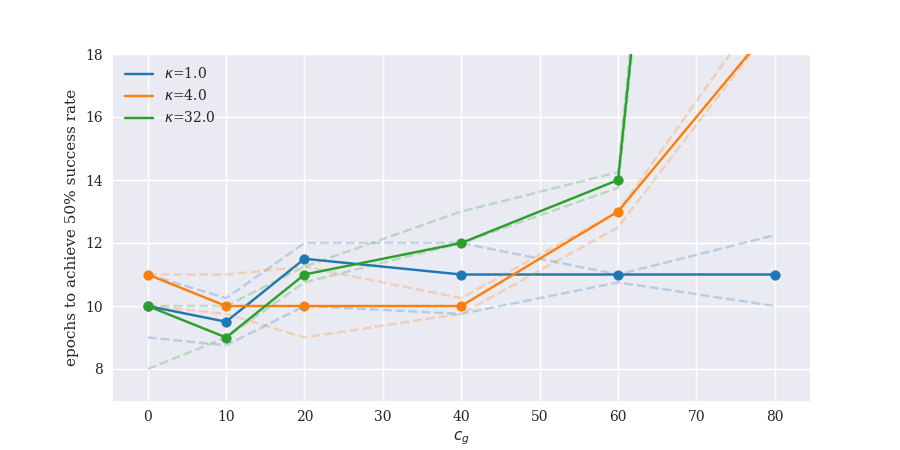}
\hspace{-2.7cm}\begin{minipage}[b][26pt][t]{2.4cm}\tiny (d) Pick-and-place DDPG+HER\end{minipage} 
\caption{Number of epochs to achieve 50\% success rate. Left: push object task, right: pick-and-place task, top: with HER, bottom: without HER.  We display the median and the quartiles (dashed lines) for $n \geq 5$ runs for each data point. }
\vspace{-8pt}
\label{fig:gr_c_graph}
\end{figure*}

A problem is that the number of masks, i.e., subgoal combinations, grows exponentially with the number of goal dimensions, which renders a naive version of the masking approach feasible only for low-dimensional goals. To overcome this issue, we assume that the success of achieving each individual subgoal is conditionally independent of the success of achieving other subgoals. With this assumption, we only need to record the successes of each individual subgoal during the evaluation phase of each epoch, and we approximate the success chance $c_m$ of a goal mask $m$ as the product of the success chances of the individual subgoals that are not masked.

This allows us to address our central research question, because we can now investigate whether there exists an optimal estimated success chance $c_g$, such that the convergence speed is maximal if we simplify goals by applying goal masks that have a success rate $c_m$ close to $c_g$. Formally, we implement the goal mask sampling by selecting a mask $m \in M$ with a probability $p_g \propto |c_{m} - c_g|^\kappa$. The exponent $\kappa$ controls the sharpness of the probability distribution.

\section{EXPERIMENTS AND RESULTS}
\label{sec:exp_results}
We evaluate our approach by running two goal-based RL algorithms on two different benchmarking problems for various parameterizations. 
We use DDPG with and without HER in a robotic simulation of a fetch robot, depicted in \autoref{fig:experiments}, performing different object manipulation tasks. We train the agents with a Gaussian action noise with $\sigma=0.2$ and exploration rate of $\epsilon=0.3$. For the HER agent, we use the \emph{future} strategy \cite{Andrychowicz2017} with a hindsight rate of $k=6$, i.e, for six HER-modified state transitions in the replay buffer one is not modified. During each epoch, 4 parallel rollouts and 64 rollout cycles are performed. To estimate the success chance of the individual subgoals, we compute the average over the last $h=10$ evaluation rollouts. 

The action space of the robot involves movement of the end effector in the x/y/z axes, realized with the inverse kinematics spring model of the MuJoCo Physics engine as well as a scalar value representing the grasping state of the robot's gripper. The observation space involves the center coordinates of the block and the end effector, their velocities, and also the end effector's grasping state.

\subsection{Experiments}

\noindent \textbf{Push the object to a position on the table.} \quad

The robot arm has to push an object to a random location on the table's surface. 
Since lifting the object is not required, the robot does not have to use its gripper.

\medskip
\noindent \textbf{Pick up and place an object at a position above the table.} \quad

The robot has to pick up the object and move it to a goal position above the table's surface. The task can be expressed as a sequence of three individual causally related subtasks that the robot has to learn implicitly, namely, moving the gripper to the object, grasping the object, and moving the object to the target location.

\begin{figure*}
\centering
\includegraphics[width=.94\textwidth,trim={0.45cm 0.0cm 2.5cm 0.3cm},clip]{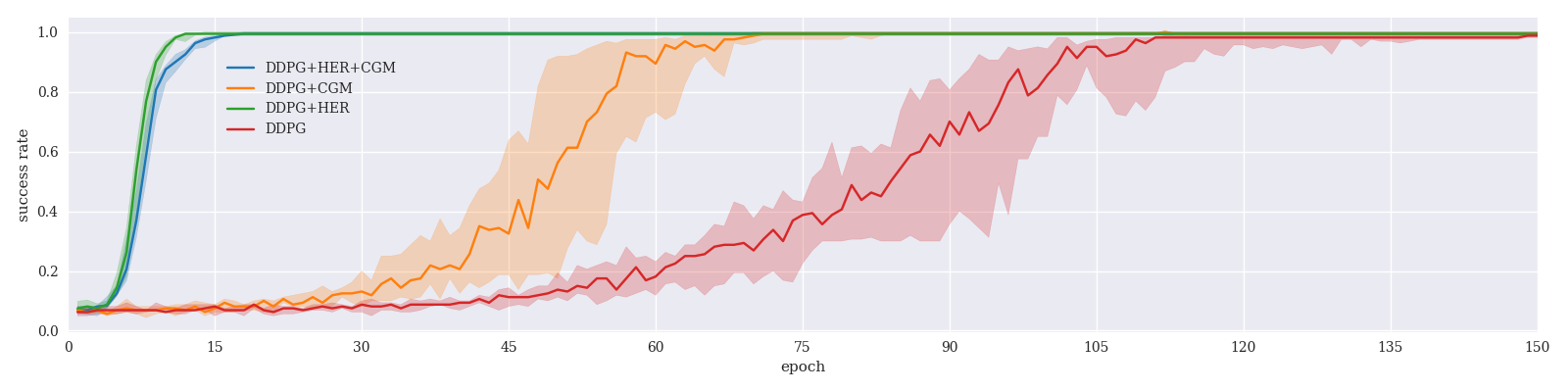}
\hspace{-2.5cm}\begin{minipage}[b][30pt][t]{2.0cm}\tiny (a) Push object\end{minipage} 
\includegraphics[width=.94\textwidth,trim={0.45cm 0.0cm 2.5cm 0.3cm},clip]{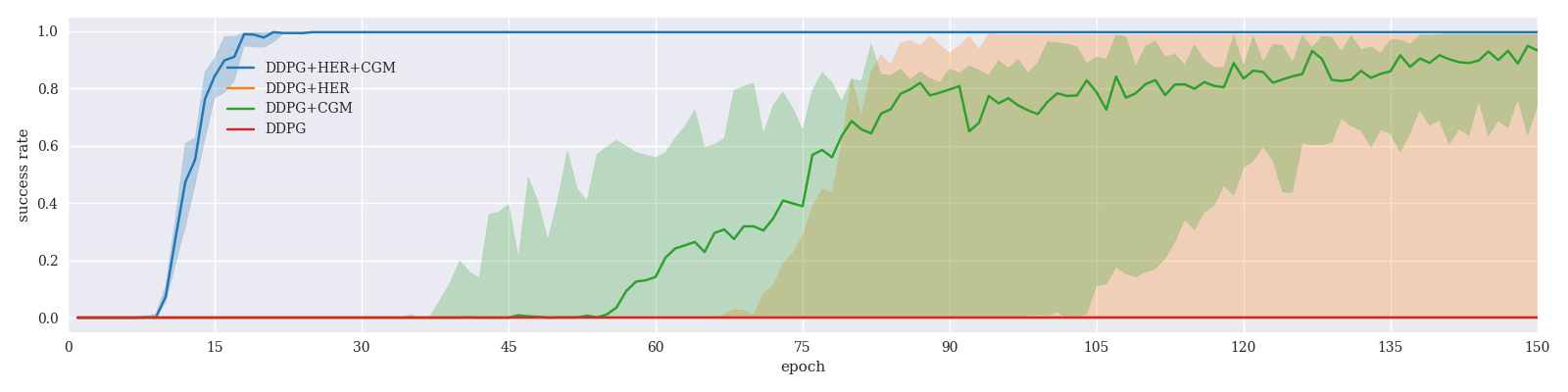}
\hspace{-2.5cm}\begin{minipage}[b][30pt][t]{2.0cm}\tiny (b) Pick and place\end{minipage} 
\vspace{-.5cm}
\caption{Learning progress for CGM compared to not using CGM. The figures show the median and the area between the upper and lower quartile over $n \geq 5$ runs for each configuration. For (b) pick and place, DDPG+HER and DDPG both had a median success rate of 0.0.}
\label{fig:conv_graph}
\end{figure*}


\subsection{Results}
\label{sec:results}
\noindent \textbf{What is the optimal ``Goldilocks'' difficulty level?} \quad

\autoref{fig:gr_c_graph} illustrates the results for the pick-and-place and the pushing experiment with a single object.
For both experiments, we run tests for different target success rates $c_g \in \{0,10,20,40,60,80\}$ and values for $\kappa \in \{1,4,32\}$, where $\kappa=1$ causes a relatively homogeneous sampling of goal masks compared to $\kappa=32$, which causes the sampling to follow a very sharp sampling distribution. 
The graphs show at which epoch an evaluation success rate of 50\% is achieved for different values for $c_g$ and $\kappa$. For the HER-based algorithm, the learning performance decreases significantly for success chances above $c_g=60\%$. It is roughly constant for values below 40\%, with a minimum at $c_g=10\%, \kappa=32$ for the pick-and-place task and $c_g=40\%, \kappa=4$ for the pushing task. 
For DDPG without HER, the minima are at $c_g=40, \kappa=32$ for the pushing task and $c_g=40, \kappa=4$ for the pick-and-place task. 

To find these parameters, we executed more than 5.000.000 training and evaluation rollouts in total. The number of used CPUs has an effect on the interleaving of rollout execution and neural network training because each CPU independently performs parallel rollouts which are merged during the experience replay to perform the training. Since we use 16 CPUs for the pick-and-place task and 4 CPUs for the pushing task, the results between both experiments are not comparable. For example, DDPG requires less epochs for the pick-and-place task with 16 CPUs than for the pushing task with 4 CPUs.

\vspace{15pt}
\noindent \textbf{Improvement of the learning performance.} \quad

To evaluate whether the CGM approach outperforms state-of-the-art algorithms that do not use goal mask sampling, we select the best values $c_g$ and $c$ for each experiment, as identified in \autoref{fig:gr_c_graph}, and compare CGM to the case where no goal mask sampling is applied. 
As we observe in \autoref{fig:conv_graph}, goal masking has a significant positive effect for all cases but the pushing task with DDPG+HER, where we observe a slightly inferior performance when using CGM. However, for the pushing task, there is an improvement when combining CGM with DDPG only: convergence is reached after approximately 70 epochs with CGM, compared to approximately 110 epochs for the case without CGM.  
\begin{figure*}[ht!]
\centering
\includegraphics[width=.47\textwidth,trim={1.38cm 0.1cm 2.5cm 1.3cm},clip]{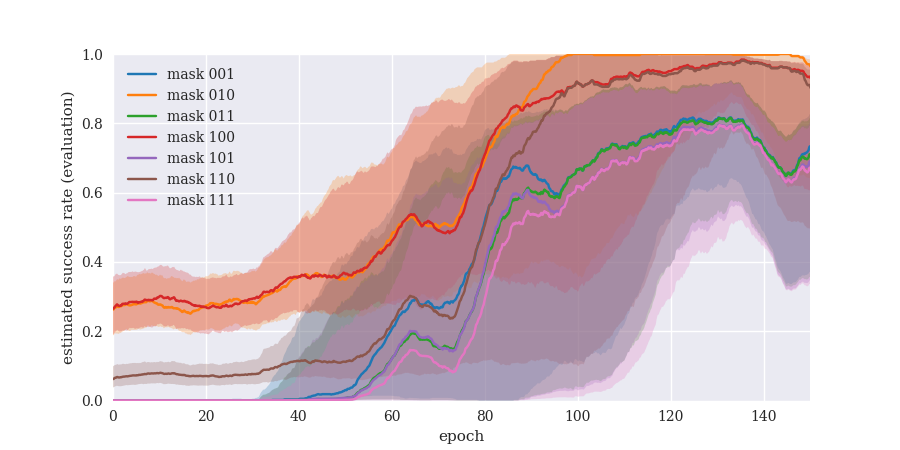}
\hspace{-1.0cm}\begin{minipage}[b][1.8cm][t]{0.75cm}\scriptsize(a)\end{minipage} 
\includegraphics[width=.47\textwidth,trim={1.38cm 0.1cm 2.5cm 1.3cm},clip]{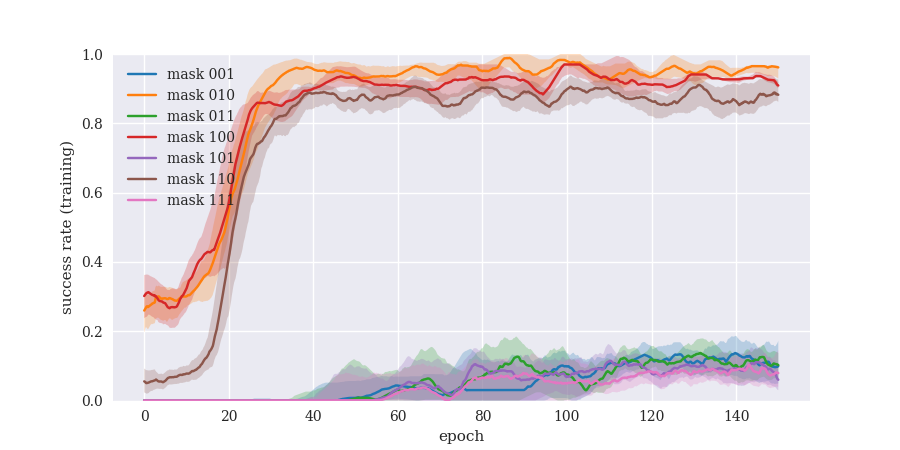}
\hspace{-1.0cm}\begin{minipage}[b][1.8cm][t]{0.75cm}\scriptsize(b)\end{minipage} 
\vspace{-0.3cm}
\caption{Goal mask specific success rates over time for the pick-and-place task with DDPG. (a) estimated success rate under the conditional independence assumption. (b) actual success rate during training rollouts. We display the median and the area between the upper and lower quartile for each mask over $n \geq 5$ runs. Note that the masks are multiplicative, i.e. mask 111 means no change in input.}
\label{fig:succ_rates_over_time}
\vspace{-0.2cm}
\end{figure*}
The largest performance gain is observed for the pick-and-place task, where DDPG alone has never been able to achieve the goal after 300 epochs.
If DDPG is combined with HER, convergence is reached only in the upper quartile (orange area in \autoref{fig:conv_graph} (b)) of all test runs. In the median, DDPG+HER was not able to reach the pick-and-place target.\footnote{Note that our results are not comparable to the original results by \citet{Andrychowicz2017} because in their work, the authors add demonstration samples to improve the learning performance.} 
If DDPG and DDPG+HER are combined with CGM, they are consistently able to reach convergence for the pick-and-place task. The median for DDPG+CGM reaches convergence after approximately 170 epochs, and the median for DDPG+HER+CGM reaches convergence after approximately 20 epochs. 
Hence, we conclude that, for the pick-and-place task, the effect on the increase of the learning performance of CGM on the DDPG algorithm is larger than the effect of HER on DDPG, and that the combination of HER with CGM synergizes well.

\medskip
\noindent \textbf{Generalizability of the approach.} \quad

To investigate whether the increase of the learning performance caused by CGM generalizes also over other reinforcement learning tasks, we investigate whether our core assumption, pertaining to the conditional independence of the probability of achieving individual subgoals, holds. To this end, we compute the estimated success chance for each goal mask for the pick-and-place task and compare it to the training success chance for that goal mask. The results are illustrated in \autoref{fig:succ_rates_over_time}. 
%
%
%
%
%
%
The training success rate for the goal masks that do not involve the masking of the z-axis is significantly lower than the estimated success rates for those masks. The reason for this difference is that during training we apply random actions using the $\epsilon$-greedy strategy, and also an additional noise on actions, which causes around 90\% of all training rollouts not to succeed for masks that do not involve the z-axis. 
However, \autoref{fig:succ_rates_over_time} also depicts that the relative increase of the training success rates corresponds to the relative increase of the estimated success rates, especially for the interesting cases where the z-axis is not masked. Here, the success chance grows after approximately 40 epochs, passes a local minimum at approximately 75 epochs, and reaches convergence at around 150 epochs in both \autoref{fig:succ_rates_over_time} (a) and (b). 
Furthermore, the relative variance of the estimated success rates over all runs is similar to the relative variance of the training success, especially for masks that do not involve the z-axis. We have observed similar results for the case of using DDPG+HER. 

We conclude that the conditional independence assumption is appropriate for the investigated tasks, and that the approach is generalizable over similar problems where the success chance distribution over individual subgoal combinations is not homogeneous.

\section{CONCLUSIONS}
\label{sec:conclusion}
We propose curriculum goal masking (CGM) as a general add-in to substitute uniform random goal sampling for any continuous deep RL algorithm that uses goal-independent policies or value functions. We have investigated the effect of CGM on DDPG with and without HER for two different object manipulation tasks. The results indicate that CGM significantly improves the learning performance, especially for the more difficult pick-and-place task. 


We show that the conditional independence assumption which underlies the approach is appropriate for the investigated tasks, as it allows us to identify the optimal ``Golidlocks'' conditions under which the learning performance is maximal. We quantify the optimal estimated success chance $c_g$ of a sampled goal to lie between a range of 10\% to 40\%, depending on the experience replay mechanism and the specific task to solve. This result coincides with \citeauthor{Florensa2018AutomaticAgents}'s observation that the range for the estimated success change of subgoals, which they refer to as GOID range, has a relatively low effect on the learning performance as long as it remains within reasonable bounds \cite{Florensa2018AutomaticAgents}.

Our results indicate the appropriateness of the approach for tasks like the pick-and-place problem, where metric coordinates represent subgoals and where the conditional independence assumption is appropriate.
Future research directions involve the investigation of more complex tasks, such as problems that involve tool use. 





\bibliographystyle{IEEEtranN}
\bibliography{paper}{}

\end{document}